# Intent Detection for code-mix utterances in task oriented dialogue systems

Pratik Jayarao
Research And Development
Haptik Infotech Pvt Ltd
Mumbai, India
pratik.jayarao@haptik.co

Aman Srivastava
Research And Development
Haptik Infotech Pvt Ltd
Mumbai, India
aman.srivastava@haptik.co

*Abstract*— Intent detection is an essential component of task oriented dialogue systems. Over the years, extensive research has been conducted resulting in many state of the art models directed towards resolving user's intents in dialogue. A variety of vector representation for user utterances have been explored for the same. However, these models and vectorization approaches have more so been evaluated in a single language environment. Dialogue systems generally have to deal with queries in different languages and most importantly Code-Mix form of writing. Since Code-Mix texts are not bounded by a formal structure they are difficult to handle. We thus conduct experiments across combinations of models and various vector representations for Code-Mix as well as multi-language utterances and evaluate how these models scale to a multi-language environment. Our aim is to find the best suitable combination of vector representation and models for the process of intent detection for code-mix utterances. We have evaluated the experiments on two different dataset consisting of only Code-Mix utterances and the other dataset consisting of English, Hindi, and Code-Mix( English-Hindi) utterances.

Keywords— dialogue, machine learning, nlp

## I. INTRODUCTION

In the recent years the number of human-bot systems driven by either voice or text have increased exponentially. Intent detection forms an integral component of such dialogue systems. We define a dialogue system where a user directs a query at the system and the system classifies the query into a given intent. Example: User: Give the novel The Secret a 4 out of 6. System (Intent): RateBook However the biggest challenge faced by these systems is that a large part of the user utterances stem from different languages and many follow the Code-Mix form of writing. A system has to be capable of handling the above forms of queries. Code-mix data posts a wide variety of challenges. It is not dictated by any formal form of grammar making it difficult to extract a pattern from the utterances. The words have no standard spellings thus leading to a wide variety of spellings for the same word in a given context. Slangs, inconsistencies in sentence formation, incorporation of non standard vocabulary make it difficult for a system to understand such utterances. We conduct experiments across two critical components of the intent detection system, Models and Vector Representations. We incorporate a wide variety of traditional models in our experiments ranging from Logistic Regression, Support.

Vector Machines, K Nearest Neighbors, Random Forest, Decision Tree. In the past few years, end to end neural network architectures have shown great promise. Thus we implement Neural Networks inspired models alongside their variations such as Deep Neural Networks, Recurrent Neural Networks, GRUs, LSTMs. We represent the user utterances with the help of various vectorization approaches. We include the more traditional, CountVectorizer and TFIDFVectorizer. We furthermore use the more recent embeddings based approaches using SGNS, USE, ELMO models to obtain the vectors for the utterances. We thus in our experiments explore different combinations of models and vector representation examining the ability of our system to handle multi-language and Code-Mix utterances.

## II. RELATED WORK

Intent detection has been a very popular problem for dialogue systems. Initial work on intent detection has been focused on supervised learning methods. Almost all standard approaches to classification have been applied in intent classification. [1] used a combination of linear support vector machines and hidden markov models for intent classification. In addition to this Bayesian based methods [2] were one of the popular approaches for the task of intent classification [3], [4]. Other traditional approaches such as use standard Decision Tree Classifier were implemented to solve the intent detection problem [5]. A wide range of rule-based approaches also have been successfully implemented to tackle the intent detection issue in dialogue systems [6]. Deep learning approaches are not dependent on hand engineered features. Instead, they have the ability to generate features at word level [7]. Experiments have been conducted where Deep Learning networks have constructed features from sentences [8], [9] and even long texts [10]–[12]. [13] incorporated word embeddings while [14] implemented intent detection using semantically enriched word embeddings. With recent developments in deep neural networks, intent detection models [15]–[18] are used to categorize intents successfully given varied utterances. Experiments on open-domain conversations have been conducted by [8], [19], [20]. [19] CNNs were used to extract local features from each utterance and RNNs were used to create a general view of the whole dialogue. [21] applied a deep LSTM structure to obtain state of the art results on Switchboard and MRDA corpuses. However all these works have been largely performed on English corpuses. We try to evaluate the capabilities of some of these approaches on a multi-language and Code-Mix utterances.

## III. DATASET

We primarily use data from from publicly available Snips 2017-06-custom-intent-engines dataset. The dataset comprised of a well balanced 7 intents SearchCreativeWork, GetWeather, BookRestaurant, PlayMusic, AddToPlaylist, RateBook, SearchScreeningEvent. As our focus was to conduct experiments across English, Hindi and Code Mix (English-Hindi) data we constructed two datasets. First, we converted the utterances into Code-Mix EnglishHindi utterances with the help of content writers Dataset1. We split this dataset into 7000 training samples and 700 test samples with each intent having equal contribution in both the training as well as the test set. Second, we translated the English data into Hindi using Google translate [22]. We then combined this translated utterances with the original English utterances and the Code-Mix utterances to form Dataset2 consisting of all three English, Hindi and Code-Mix data. The Dataset2 was split into 21000 training samples and 2100 test samples. We evaluated the results using macro F1-Score.

## IV. ENCODERS

### A. Word Embeddings

Word Embeddings have a significant role in the increasing the performance of models as they exploit the syntactic and semantic understanding of a word. We conducted experiments with two different (frozen) embeddings w.r.t. dimension size.

### B. Skip Gram Negative Sampling (SG25, SG100)

We trained 25, 512 dimensional word embeddings using the SGNS [23] model. We restricted the training data to training set for each experiment.

### C. ELMO

We incorporated the Elmo [24] model in our experiments. This model provides a 512 dimensional representation for each word. The model constructs contextualized word representations using character-based word representations and bidirectional LSTMs.

### D. Universal Sentence Encoder (USE)

The above stated approaches work on a word-level encoding. We additionally performed experiments using the Universal Sentence Encoder [25] which has the capability to encode a sentence to fixed size of 512 dimensional embeddings.

## V. VECTORS

### A. Count Vectorizer (Count)

The CountVectorizer is used for tokenizing a set of documents and generate a vocabulary of known words. The vectorizer encodes an utterance and returns a vector with the length equal to the size of the entire vocabulary and an integer count for the number of occurrences for each token.

### B. TFIDF Vectorizer (Tfidf)

TF-IDF where TF stands for term frequency and IDF stands for inverse document score provides token frequency scores that highlight words frequent in a document but not across documents. The TfidfVectorizer tokenizes documents, learns the vocabulary and inverse document frequency weightings, which are then used to encode new utterances.

### C. Latent Semantic Analysis (Lsa)

LSA takes the tfidf vectorizer a step further, it decomposes it into further separate document-topic matrix and a topic-term matrix using SVD (Singular Value Decomposition). This helps to reduce the feature set of each documents and tries to map a document with some concept (topic), which is not possible in count and tfidf vectorizer.

### D. Word Embeddings Average (Avg)

The above vectors however fail to provide any semantic knowledge to our models. We performed an unweighted average of all the tokens present in a given message to obtain a fixed size vector for each message.

### E. Word Embeddings (Idf Avg)

We performed an idf weighted average of all the tokens present in a given message to obtain a fixed size vector for each message. Thus we were able to incorporate the semantic knowledge as well as weight contributed by idf score of the token.

## VI. MODELS

### A. Support Vector Machines (SVM)

A Support Vector Machine performs categorization by obtaining the hyperplane that maximizes the margin between the classes. The vectors that form the hyperplane are the known as the support vectors. This hyperplane is then further used to predict classes for new instances.

### B. Logistic Regression Classifier

Logistic Regression is a supervised model and is used where the output variable is categorical. The idea of logistic regression is to find conditional probability, of output given its input. Logistic regression uses Logit Functions or log-odds function, for calculation of conditional probability.

### C. K Neighbors Classifier

The K-Nearest Neighbors (K-NN) is a algorithm utilized for classification and regression and is non-parametric in nature. Predictions process for a new instance is to search through the entire training data for the K most similar instances or neighbors and encapsulate the output variable for these specific instances.

### D. Random Forest Classifier

Random forests or random decision forests are an ensemble learning method for classification, regression etc. They function by generating a large number of decision trees at training and outputting the class that is the mode of the classes (classification) or mean prediction (regression) of the individual trees... Random Forest are used to tackle the decision forests nature of overfitting to their training set.

### E. Decision Tree Classifier

The decision tree classifiers constructs a sequence of conditions in a tree structure. The root and internal nodes contain attribute test conditions and terminal nodes are assigned a class labels Algorithms for constructing decision trees usually work top-down, by choosing a variable at each step that best splits the set of items. Algorithms to generate decision trees work by selecting a variable at each step. Approaches such as Information gain, Gini-Index are used to find the best fit variable for the same.

TABLE I. MACRO AVERAGED F1 SCORES ON DATASET1

| Classifiers | Count | Tfidf | Count-Lsa | Tfidf-Lsa | SG25-Avg | SG25-Idf Avg | SG-512 Avg | SG-512 Idf Avg | USE | ELMO |
|---|---|---|---|---|---|---|---|---|---|---|
| Linear SVM | 91.18 | 91.43 | 91.16 | 91.17 | 93.58 | 94.15 | 92.50 | 94.82 | 91.01 | 84.67 |
| Logistic Regression | 91.60 | 91.30 | 91.20 | 92.49 | 92.72 | 94.73 | 91.42 | 94.17 | 89.25 | 84.01 |
| K Neighbors | 87.42 | 89.91 | 89.72 | 90.28 | 93.16 | 92.80 | 90.88 | 91.76 | 86.32 | 87.52 |
| Random Forest | 91.28 | 91.56 | 87.71 | 90.15 | 93.56 | 93.64 | 91.44 | 93.80 | 83.93 | 84.65 |
| SVM | 84.61 | 86.49 | 88.71 | 84.14 | 91.66 | 91.91 | 92.52 | 92.93 | 85.47 | 85.16 |
| Neural Networks | 93.58 | 93.67 | 93.74 | 93.58 | 94.53 | 95.01 | 94.93 | 95.16 | 91.16 | 84.88 |
| Cosine Similarity | 92.83 | 85.01 | 90.58 | 91.18 | 95.42 | 93.90 | 94.18 | 93.44 | 85.97 | 93.20 |
| Decision Tree | 90.88 | 90.89 | 89.21 | 89.68 | 91.96 | 91.88 | 91.46 | 92.70 | 82.63 | 82.55 |

TABLE II. MACRO AVERAGED F1 SCORES ON DATASET2

| Classifiers | Count | Tfidf | Count-Lsa | Tfidf-Lsa | SG25-Avg | SG25-Idf Avg | SG-512 Avg | SG-512 Idf Avg | USE | ELMO |
|---|---|---|---|---|---|---|---|---|---|---|
| Linear SVM | 89.50 | 89.76 | 85.84 | 86.31 | 92.96 | 77.10 | 93.43 | 78.24 | 84.72 | 94.70 |
| Logistic Regression | 89.26 | 88.85 | 85.27 | 84.53 | 93.04 | 77.28 | 92.63 | 77.78 | 78.86 | 93.05 |
| K Neighbors | 84.35 | 81.88 | 81.88 | 84.21 | 93.95 | 76.97 | 93.29 | 77.48 | 83.88 | 89.18 |
| Random Forest | 86.17 | 86.84 | 82.77 | 83.79 | 92.33 | 78.23 | 92.24 | 78.29 | 75.75 | 75.18 |
| SVM | 80.68 | 80.04 | 80.55 | 79.47 | 93.59 | 78.49 | 84.60 | 75.51 | 74.42 | 74.80 |
| Neural Networks | 89.98 | 90.83 | 87.90 | 88.27 | 94.27 | 79.93 | 94.98 | 79.59 | 87.66 | 94.23 |
| Cosine Similarity | 84.82 | 84.97 | 83.17 | 84.42 | 94.19 | 76.60 | 93.85 | 76.11 | 84.16 | 89.53 |
| Decision Tree | 83.27 | 84.36 | 81.67 | 82.35 | 91.93 | 77.97 | 91.64 | 75.96 | 74.53 | 74.00 |

TABLE III. MACRO AVERAGED F1 SCORES FOR RECURRENT NEURAL MODELS ON DATASET1

| Classifiers | SG25 | SG512 | ELMO |
|---|---|---|---|
| RNN | 95.65 | 95.86 | 95.73 |
| GRU | 96.28 | 97.01 | 96.96 |
| LSTM | 95.72 | 96.86 | 96.31 |

TABLE IV. MACRO AVERAGED F1 SCORES FOR RECURRENT NEURAL MODELS ON DATASET2

| Classifiers | SG25 | SG512 | ELMO |
|---|---|---|---|
| RNN | 95.05 | 95.13 | 95.10 |
| GRU | 95.48 | 95.64 | 95.51 |
| LSTM | 95.32 | 96.54 | 95.44 |

*F. Neural Networks*

A feedforward artificial neural network model that has one or more of hidden layer units which are encapsulated by nonlinear activations. It utilizes backpropagation for training. Here Learning is achieved in the neuron by altering connection weights after data is processed, based on the difference in the output compared to the expected result.

*G. Cosine Similarity*

The cosine similarity between two vectors is a measure that calculates the cosine of the angle between them. For a new instance of data we compare the cosine similarity between the new instance and the instances present in the training set. The training instance which has the maximum similarity score with new instance is selected and the corresponding label is assigned to the new instance.

*H. Recurrent Neural Network (RNN)*

A recurrent neural network (RNN) is a class of artificial neural network where connections between units form a directed graph along a sequence. This allows it to exhibit dynamic temporal behavior for a time sequence. RNN [22], [23] consists of a hidden state that depending on the previous hidden state and current in-put continually updates itself at every time step. The output is then predicted on the basis of the new hidden state.

*I. Gated Recurrent Units (GRU)*

To solve the vanishing/exploding gradient problem this GRU[26] uses update gate and reset gate. These vectors decide the information that should be passed to the output. They can trained to retain long dependencies or discard information which is irrelevant to the prediction.

*J. Long Short Term Memory (LSTM)*

Long Short Term Memory [27] networks are the same as RNNs, except that the hidden layer updates are replaced by purpose-built memory cells. As a result, they may be better at finding and exploiting long range dependencies in the data.

VII. EXPERIMENTS

We conducted an wide range of experiments across Dataset1 (Code-Mix) and Dataset2 (English-Hindi-CodeMix). The results are reported in TABLE I, TABLE II, TABLE III and TABLE IV. Our focus was to evaluate the best model and vector settings suited for intent detection. Thus we experimented with various combinations of models and vectors. We have summarized the results of the experiments below.

*A. Implementation*

We implemented LinearSvc, Logisticregression, Kneighborsclassifier, RandomForest, DecisionTree SupportVectorMachine, Cosine Similarity, CountVectorizer,

TfidfVectorizer, LSA using the Scikit-learn library [28]. The Neural Networks, RNN, GRU, LSTM architectures were consisted of two hidden layers. Early stopping was implemented to handle overfitting. We implemented neural network based architectures using Tensorflow. [29]. We incorporated the USE and ELMO pre-trained models available on Tensorflow Hub.

*B. Choice of Architecture*

The non-neural network based models displayed promising results. The Neural Network model displayed a higher score compared to these traditional models. However, RNN, GRU and LSTM based architectures displayed the best performance as they were able to incorporate the temporal component of the text utterances. Through the course of our experiments GRU based network displayed the highest F1-Score.

*C. Choice of Vector Representation*

The Count and Tfidf based representations along with their Lsa representations displayed encouraging results but were unable to include any form of semantic or contextual information for the given utterances. The encoder based representations for utterances (SG25, SG512, USE, ELMO) provide higher accuracy as they are able to incorporate semantic knowledge into the vector representations with self trained SG512 displaying the highest F1-Score on average. The highest F1-Score was displayed by GRU based Neural Network along with SG512.

VIII. CONCLUSION

Thus we were able to successfully explore various combinations between the models and vector representations for Code-Mix and multi-language utterances. Traditional approaches provided encouraging results. However, Deep Recurrent Neural Networks displayed a significant increase in performance compared to the aforementioned models. In terms of vector representation the more traditionally used approaches were outperformed by high dimensional self-trained embeddings which were able to incorporate semantic meanings into the vector representations. Thus we can conclude that a combination of Deep Recurrent based Networks and self-trained embeddings are able to scale to Code-Mix and multi-language utterances seamlessly.